\documentclass[runningheads]{llncs}
\usepackage[T1]{fontenc}
\usepackage{graphicx}
\usepackage{amsmath}
\usepackage{booktabs}
\usepackage{caption}
\usepackage{subcaption}
\usepackage{multirow}

\usepackage[misc]{ifsym}
\usepackage{lipsum}

\begin{document}
%
\title{Attentive Symmetric Autoencoder for Brain MRI Segmentation}
%
%

\author{Junjia Huang\inst{1,2}
\and
Haofeng Li\inst{1}$^{(\textrm{\Letter})}$
\and
Guanbin Li\inst{1,2}$^{(\textrm{\Letter})}$
\and
Xiang Wan\inst{1,3}
}

%
\authorrunning{J. Huang et al.}
%

\institute{Shenzhen Research Institute of Big Data, The Chinese University of Hong Kong, Shenzhen, China\\
\and
School of Computer Science and Engineering, Sun Yat-sen University, Guangzhou, China
\and
Pazhou Lab, Guangzhou, 510330, China\\
\email{lhaof@foxmail.com, liguanbin@mail.sysu.edu.cn}
}

%
\maketitle              

\newcommand\blfootnote[1]{%
\begingroup
\renewcommand\thefootnote{}\footnote{#1}%
\addtocounter{footnote}{-4}%
\endgroup
}

\begin{abstract}
Self-supervised learning methods based on image patch reconstruction have witnessed great success in training auto-encoders, whose pre-trained weights can be transferred to fine-tune other downstream tasks of image understanding. However, existing methods seldom study the various importance of reconstructed patches and the symmetry of anatomical structures, when they are applied to 3D medical images. In this paper we propose a novel Attentive Symmetric Auto-encoder (ASA) based on Vision Transformer (ViT) for 3D brain MRI segmentation tasks. We conjecture that forcing the auto-encoder to recover informative image regions can harvest more discriminative representations, than to recover smooth image patches. Then we adopt a gradient based metric to estimate the importance of each image patch. In the pre-training stage, the proposed auto-encoder pays more attention to reconstruct the informative patches according to the gradient metrics. Moreover, we resort to the prior of brain structures and develop a Symmetric Position Encoding (SPE) method to better exploit the correlations between long-range but spatially symmetric regions to obtain effective features. Experimental results show that our proposed attentive symmetric auto-encoder outperforms the state-of-the-art self-supervised learning methods and medical image segmentation models on three brain MRI segmentation benchmarks.
\blfootnote{This work is supported in part by Chinese Key-Area Research and Development Program of Guangdong Province (2020B0101350001), in part by the National Natural Science Foundation of China (No.62102267), in part by the Guangdong Basic and Applied Basic Research Foundation (No.2020B1515020048), in part by the National Natural Science Foundation of China (No.61976250), in part by the Guangzhou Science and technology project (No.202102020633), and the Guangdong Provincial Key Laboratory of Big Data Computing, The Chinese University of Hong Kong, Shenzhen. Haofeng Li and Guanbin Li are corresponding authors.}

\end{abstract}
\section{Introduction}
Accurate segmentation of brain lesion, tumour or tissue for Magnetic Resonance Imaging (MRI) data is essential for building a computer-aided diagnosis (CAD) system, and helps medical experts improve diagnosis and treatment planning. It is necessary to develop an automatic segmentation tool for brain MRI.

Deep convolutional neural networks (DCNNs) have achieved success in brain MRI segmentation~\cite{cciccek20163d,ronneberger2015u,zhou2018unet++}, but their local receptive fields fail to capture long-range spatial dependencies. Recently, transformer-based models~\cite{dosovitskiy2020image,li2022view} have drawn extensive attention and shown the state-of-the-art results on 3D image segmentation~\cite{wang2021transbts,hatamizadeh2022unetr,zhou2021nnformer}.
These methods collect dense correlations between long-range voxels for representation learning, but they require numerous voxel-level annotations that is scarce in brain medical image.
Self-supervised learning (SSL) \cite{taleb2021multimodal,taleb20203d,tao2020revisiting} uses unlabeled data to pre-train a model that can be fine-tuned to improve the results on downstream tasks. Recently, reconstruction-based SSL methods~\cite{he2021masked,wei2022masked}, which pre-train transformers for patch-level recovering with natural images. 
If these methods are applied to 3D medical images, they may fail to model the prior of a brain because they treat all recovered patches equally. Some recent work~\cite{Tang_2022_CVPR} pre-trains transformers for medical images but it neglects the symmetry of brain structures and the different importance of brain regions.

Motivated by the above observations, we consider a novel transformer-based SSL framework for brain MRI segmentation. Despite individual variations, the structure of brain tissues is relatively stable while lesions have their particular textures and appearance. During the SSL, reconstructing a smooth brain region is not challenging and may cause over-fitting. On the contrary, synthesizing an informative image patch is more difficult, which requires mining the intrinsic representations of anatomical structures. In this work, we propose an attentive reconstruction loss weighting different image regions with their informativeness that is measured by a handcrafted gradient-based score. 
Moreover, symmetry is an essential prior of brain structure. As transformers encode the coordinates of image patches for computing correlations between different positions, we introduce the symmetry to design a new position encoding method which returns the same code for two distant but symmetrical positions. Transformers with the encoding can enhance the visual features by emphasizing the correlations between contralateral brain regions. Finally, we integrate the proposed loss and encoding with a masked autoencoder to build our proposed SSL framework. Our contributions are summarized as: (1) a novel attentive reconstruction loss function, (2) a new symmetric position encoding method, and (3) an SSL framework attentive symmetric autoencoder for brain MRI segmentation. (4) Experimental results show that our method outperforms the state-of-the-art SSL methods and medical image segmentation models on three public benchmarks.

\section{Methodology}

\subsection{Attentive Symmetric Autoencoder}
\begin{figure}[!t]
\includegraphics[width=\textwidth]{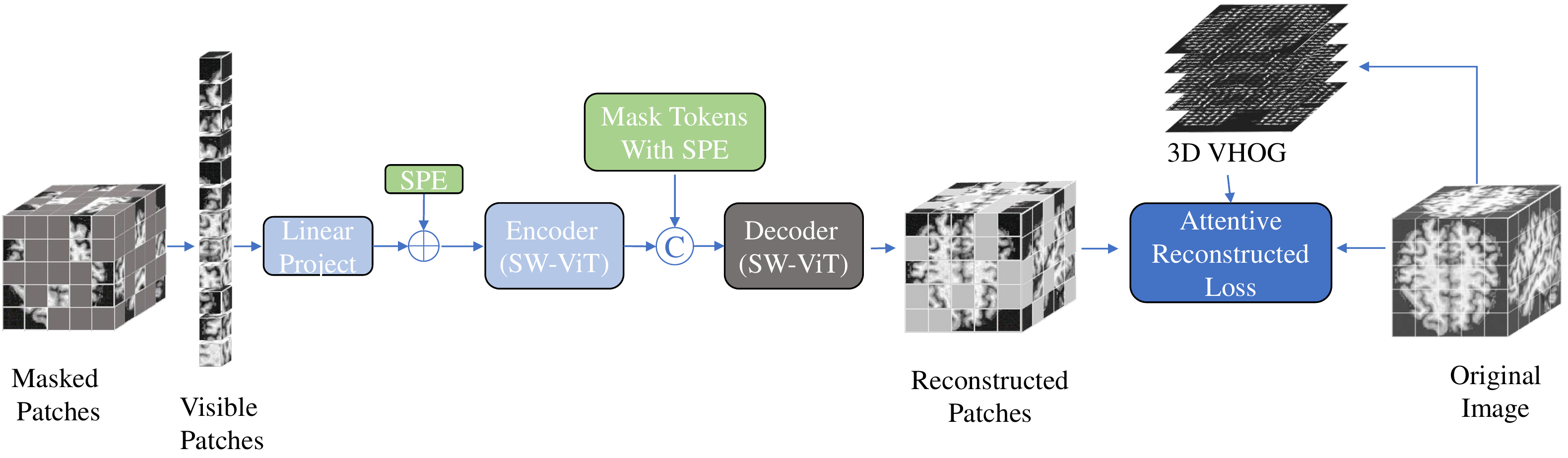}
\caption{The architecture of Attentive Symmetric Autoencoder. SPE means symmetric position encoding.} 
\label{ASA-Struct}
\end{figure}

We propose a novel Attentive Symmetric Autoencoder (ASA) that can be trained to obtain generalizable model weights for adapting brain MRI segmentation tasks. As shown in Fig \ref{ASA-Struct}, the proposed ASA consists of a pair of encoder and decoder with symmetric position encoding (SPE) and an attentive reconstruction loss. During the self-supervised training of ASA, the input 3D image is divided into regular non-overlapping image patches (of size $s\!\times\!s\!\times\!s$). $P$\% of these image patches are randomly masked and only the unmasked patches are visible. 
After a linear projection, each visible patch is embedded into a feature vector, which is added with its Symmetric Position Encoding (SPE) to produce the encoder input. The encoder outputs the same number of vectors as its input.
Mask Tokens are the same learnable vector added with different SPEs. Each mask token corresponds to a masked image patch. The encoder output is concatenated with the mask tokens to form the decoder input. The decoder reconstructs all the image patches and only the masked ones are used to compute the proposed loss.

\subsubsection{Attentive Reconstruction Loss.}
 Considering that learning to recover flatten regions is less helpful for  encouraging the model to harvest discriminative representations. We develop an attentive reconstruction loss function that emphasizes the informative regions of brain MRI. To estimate the information of an image patch, we adopt a gradient based metric for 3D images.
Inspired by 3D VHOG~\cite{hog2015}, we calculate the gradient vector $\vec{g} = (g_x, g_y, g_z)$ for each voxel by applying the filter mask of [-1, 0, 1]. In spherical coordinates, we use two scalars $\theta$ and $\phi$ to represent the orientation of a voxel. $\theta$ and $\phi$ can be calculated as: 
\begin{equation}
\begin{aligned}
\theta = cos^{-1}\Big(\frac{g_z}{\sqrt{g_x^2+g_y^2+g_z^2}}\Big), 
 \phi &= |atan2(g_y, g_x)|.
\end{aligned}
\end{equation}
For each image patch we build a 2D histogram $G$ and the number of bins is $b\times b$. To compute the values $G$, we traverse each voxel in the image patch. Let $\theta, \phi$ denote the orientation of the current voxel. We first determine the bin indexes of the voxel as $r=\lfloor \theta/(\pi/b) \rfloor, c=\lfloor \phi/(\pi/b)\rfloor$. 
And then we accumulate $||\vec{g}||$ (the gradient magnitude of the current voxel) to the corresponding bin $(r,c)$ of the 2D histogram $G$. 
After processing all voxels in an image patch, $L_2$ norm is performed on $G$, the histogram of the patch. We calculate the mean of $G$ as $\bar{G}$ for each image patch. $G$ is normalized among all $N$ masked image regions to characterize the relative importance $p_i$ as $p_i = \frac{\bar{G_i}}{\sum_{i=1}^{N}(\bar{G_i})}$.

Our proposed loss function adopts mean squared error (MSE) to measure the pixel-level difference between the recovered image areas and the original ones, and pays more attention to the informative brain regions using the gradient-based weight $p_i$. The overall loss can be formulated as Eq.~(\ref{eq_loss}):
\begin{equation}
L(X,Y) = \sum_{i=1}^N \big(p_i\cdot\sum_j^M(X_{ij}-Y_{ij})^2 /M \big).
\label{eq_loss}
\end{equation}
where $X,Y$ are the reconstructed and the original images. $N$ is the number of masked image patches in an image. $M$ is the voxel number in an image patch. $X_{ij}$ denotes the $j$-th voxel of the $i$-th patches in the image $X$.

\begin{figure}[!tb]
\includegraphics[width=\textwidth]{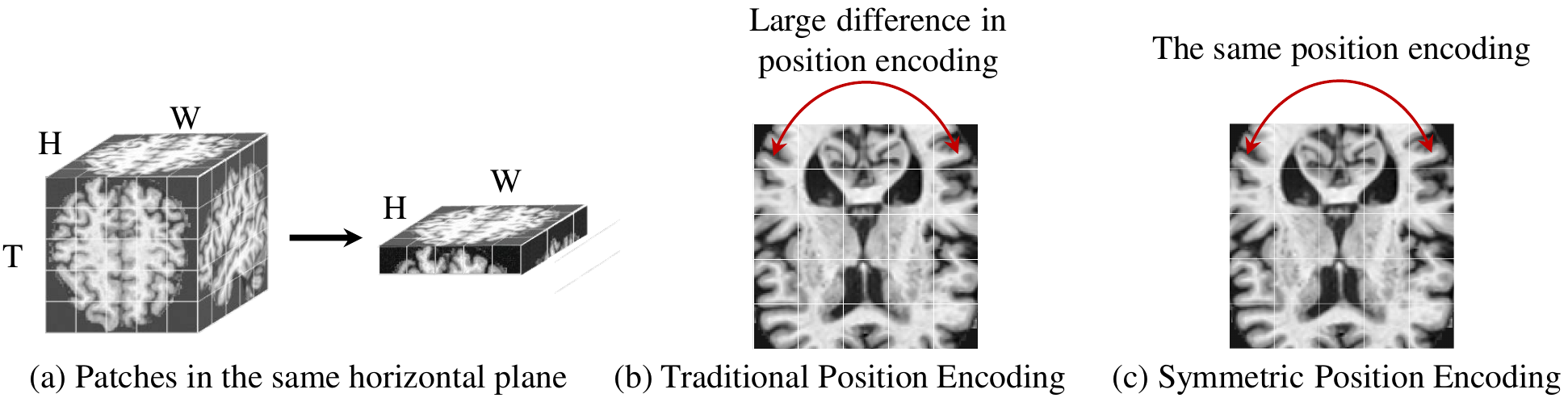}
\caption{Symmetric Position Encoding.} 
\label{SPE-structure}
\end{figure}
\subsubsection{Symmetric Position Encoding.}
We observe the left-right symmetry of brain structures, and propose a Symmetric Position Encoding (SPE) method. The proposed method narrow the encoding difference of two symmetric image positions, and can encourage the model to harvest better features from these two correlated regions.
For the patches in the same horizontal plane (Fig.~\ref{SPE-structure}(a)), the vanilla position encoding \cite{vaswani2017attention} of the top left is largely different from that of the top right (Fig.~\ref{SPE-structure}(b)), even though these regions have similar contents. However, using our proposed SPE, the leftmost and the rightmost positions (in the same row) can share the same encoding.
Let $T\!\times\!H\!\times\!W$, $(t, h, w)$ denote the patch number of an image and the coordinate of an image patch. The symmetric position encoding is computed as Eq.~(\ref{eq_spe}):
\begin{equation}
\begin{aligned}
&Pos = (T^2\cdot t+H\cdot h-|W/2-w|+W/2) / (10000^{2i/D}),\\
&P\!E(t,h,w,2i) = \sin(Pos), 
P\!E(t,h,w,2i+1)= \cos(Pos),
1\le i\le \lfloor D/2\rfloor,\\
\end{aligned}
\label{eq_spe}
\end{equation}
where $D$ is the dimension number of the SPE vector and is set to the channel number of image patch embeddings. $P\!E(\cdot)$ returns the $2i$-th/$(2i+1)$-th element of the SPE vector for a patch at $(t, h, w)$. As Fig.~\ref{ASA-Struct} shows, the SPE method is used for twice, one for patch embeddings, the other for mask tokens.

\subsection{Network Architecture}
\begin{figure}[!tb]
\includegraphics[width=\textwidth]{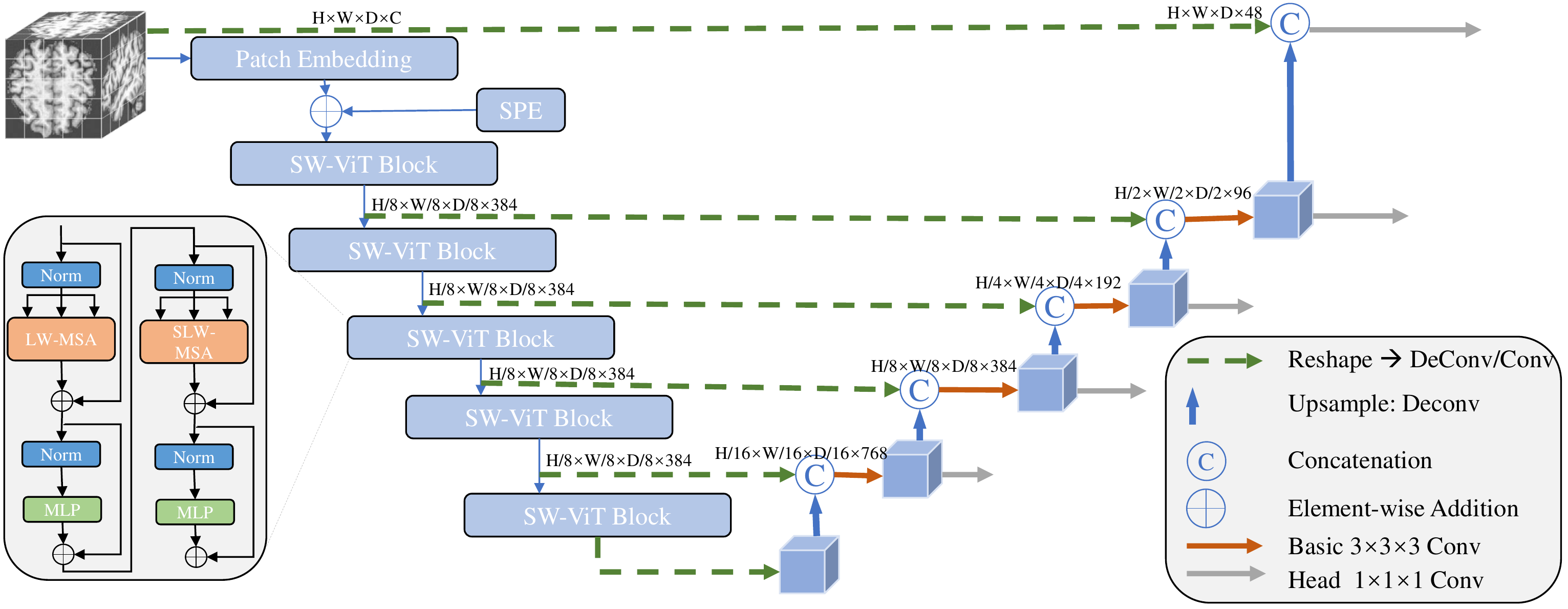}
\caption{The architecture of network in downstream tasks.} \label{Down-structure}
\end{figure}

The proposed ASA model is to provide pre-trained model weights for the downstream task, brain MRI segmentation. Here we describe the architecture of the ASA model and the image segmentation model. The encoder and the decoder of the ASA are based on Vision Transformer (ViT)~\cite{dosovitskiy2020image}. The standard ViT~\cite{dosovitskiy2020image} uses vanilla self-attention (SA), which leads to high computational cost, especially when processing 3D images. For efficiency, we develop Linear Window-based Multi-head Self-attention (LW-MSA) and Shifted Linear Window-based Multi-head Self-attention(SLW-MSA).
Inspired by SwinT~\cite{liu2021swin}, we flatten 3D patches into a sequence of patch embeddings, and split the sequence into windows of size $S$. LW-MSA computes self-attention within each 1D window. SLW-MSA shifts the sequence by $\lfloor \frac{S}{2} \rfloor$ before computing a LW-MSA module, and shifts the sequence by $\lfloor \frac{S}{2} \rfloor$ reversely after the LW-MSA module. 
LW-MSA and SLW-MSA are computed on a patch level since we convert each image patch to a feature vector via a patch embedding layer at the very beginning.
LW-MSA and SLW-MSA are stacked alternately to extract cross-windows features and to build a shifted-window ViT (SW-ViT) for our ASA model. For brain MRI segmentation, we build a U-net with the ASA encoder as the backbone, as shown in Fig.~\ref{Down-structure}.

\section{Experiments and Results}

\subsubsection{Implementation Details}
To pre-train the ASA model, we use center-cropping augmentation, Xavier uniform initializer~\cite{glorot2010understanding} for SW-ViT blocks and set the hyper-parameters following \cite{he2021masked} (see Table~\ref{table1}(a)). We follow MAE~\cite{he2021masked} and set $P$ to 75. The patch size $s$ is 8. To fine-tune the image segmentation model (Fig.\ref{Down-structure}), we adopt the online data augmentation~\cite{isensee2021nnu} (random rotation, scaling, flipping and Gamma transformation). Only the encoder of the ASA is used for initialization. Other settings are in Table~\ref{table1}(b). The experiments are run with PyTorch. For the pre-training we use four 32GB GPUs (NVIDIA V100). It takes 1 day with the early-stop strategy. The fune-tuning takes 1-2 days with 1 GPU.

\begin{table}[!htbp]
    \centering
    \caption{The hyper-parameters setting for pre-training and fine-tuning.}
    \label{table1}
    \begin{subtable}[t]{0.495\linewidth}
        \begin{tabular}{l|l}
            \ config &\ value\\
            \midrule[1pt]
            \ optimizer & \ AdamW\cite{loshchilov2017decoupled} \\
            \ optimizer momentum &\ $\beta_1,\beta_2=$0.9, 0.95\\
            \ weight decay & \ 0.05\\
            \ learning rate schedule &\ cosine decay\cite{loshchilov2016sgdr} \\
            \ warmup epochs\cite{goyal2017accurate} &\ 1e-6\\
            \ base learning rate & \ 1.5e-4\\
            \ batch size & \ 96\\
        \end{tabular}
        \caption{Pre-training setting.}
        \label{table1a}
    \end{subtable}
    \begin{subtable}[t]{0.495\linewidth}
        \begin{tabular}{l|l}
            \ config &\ value\\
            \midrule[1pt]
            \ optimizer & \ SGD \\
            \ optimizer momentum &\ 0.99\\
            \ weight decay & \ 3e-05\\
            \ initial learning rate & \ 0.01\\
            \ batch size & \ 2\\
            \ num\_epoch & \ 1000 \\
            \ loss &  Dice and CE loss \\
        \end{tabular}
        \caption{Fine-tuning setting.}
        \label{table1b}
    \end{subtable}
\end{table}

\noindent\textbf{Datasets.}
For pre-training our ASA model, we adopt T1 MRI from 2 public datasets, including 9952 cases from Alzheimer's Disease Neuroimaging Initiative(ADNI) dataset\footnote{http://adni.loni.usc.edu/} \cite{jack2008alzheimer} and 2041 cases from Open Access Series of Imaging Studies(OASIS) dataset\footnote{https://www.oasis-brains.org/} \cite{lamontagne2019oasis}. We convert the data into Brain Imaging Data Structure (BIDS), affinely align the T1 images to the MNI space via Clinica platform \cite{el2021clinica}, strip the brain skull from these images with ROBEX \cite{iglesias2011robust} and crop a $128\!\times\!128\!\times\!128$ region at their center.

For downstream task, we adopt 3 brain MRI segmentation benchmarks:
\noindent\textbf{Brain Tumor Segmentation (BraTS) 2021 dataset}\footnote{http://www.braintumorsegmentation.org/}~\cite{menze2014multimodal} has 1251 subjects. Each subject has 4 aligned MRI modalities: T1, T1Gd, T2 and T2-FLAIR. The annotations consist of GD-enhancing tumor (ET), peritumoral edematous (ED) and necrotic tumor core (NCR), which are combined into 3 nested sub-regions: Whole Tumor (WT), Tumor Core (TC), Enhancing Tumor (ET). Following \cite{zhou2021nnformer}, we set the ratio of training/validation/test as 7:1:2. 

\noindent\textbf{Internet Brain Segmentation Repository (IBSR) dataset}\footnote{https://www.nitrc.org/projects/ibsr/}~\cite{rohlfing2004evaluation} has 18 T1-weighted MRI volumes of 4 healthy females and 14 healthy males. The ground truth (GT) has 3 categories: Cerebrospinal Fluid (CSF), Gray Matter (GM), White Matter (WM). We adopt 12 cases for training and 6 cases for testing.

\noindent\textbf{White Matter Hyperintensities (WMH) dataset}\footnote{https://wmh.isi.uu.nl/data/}~\cite{kuijf2019standardized} involves 60 T1 images with pixel-level labels of White Matter Hyperintensities(WMH). We process data as \cite{li2018fully} and use 36 cases for training and the rest for testing.

\begin{table}[!t]
    \centering
    \setlength{\tabcolsep}{3mm}{
    \caption{Comparison on BraTS 2021 dataset. The first group are several competing methods. The best performance is in \textbf{bold}.}
    \label{table_brats}
    \begin{tabular}{lcccccc}
   \toprule
   Task & \multicolumn{6}{c}{BraTS 2021} \\
   Metric & \multicolumn{3}{c}{Dice(\%)$\uparrow$} &\multicolumn{3}{c}{HD95(mm)$\downarrow$}  \\
   \cmidrule(lr){2-4} \cmidrule(lr){5-7}
   Anatomy & WT & TC & ET & WT & TC & ET  \\
   \midrule[2pt]
   nnFormer \cite{zhou2021nnformer} & 91.46 & 87.42	& 82.22 & 10.15 & 9.59 & 16.78\\
   TransBTS \cite{wang2021transbts} & 92.06 & 88.20 & 79.46 & 4.98 & 4.86 & 16.32\\
   UNETR \cite{hatamizadeh2022unetr} & 92.12 & 88.32 & 79.61 & 4.91 & 4.67 & 16.32\\
   3D-RPL \cite{taleb20203d} & 93.92 & 90.13 & 85.92 & 3.74 & 3.98 & 13.71 \\
   3D-Jig \cite{taleb20203d} & 93.87 & 90.14 & 86.01 & 3.85 & 3.94 & 11.79 \\
   Ours & \textbf{94.03} & \textbf{90.29} & \textbf{86.76} & \textbf{3.61} & \textbf{3.78} & \textbf{10.25}\\
   \bottomrule
\end{tabular}}
\end{table}

\noindent\textbf{Evaluation Metric.}
We calculated Dice coefficient scores (Dice) and 95\% Hausdorff Distance (HD95) to evaluate the segmentation results in our experiments.
\subsubsection{Comparison with the State-of-the-art.} We compare our method with existing 3D transformer-based models (nnFormer~\cite{zhou2021nnformer}, TransBTS~\cite{wang2021transbts}, UNETR~\cite{hatamizadeh2022unetr}) and 3D self-supervised methods (Relative 3D patch location(3D-RPL)~\cite{taleb20203d}, 3D Jigsaw puzzle Solving (3D-Jig)~\cite{taleb20203d}) on 3 brain MRI segmentation tasks.

\begin{figure}[!t]
\includegraphics[width=\textwidth]{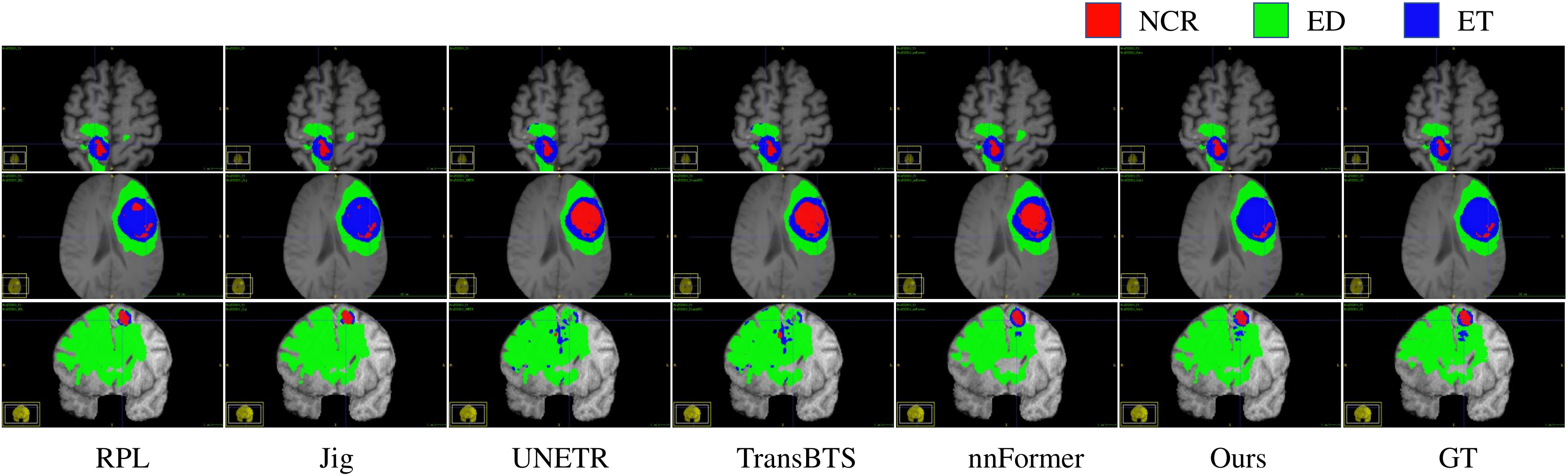}
\caption{Visualization of segmentation results on BraTS 2021 dataset.} \label{Visual_Seg}
\end{figure}
As Table \ref{table_brats} shows, on Brats 2021 dataset our method achieves the Dice scores of 94.03\%, 90.29\%, 86.76\% and the HD95 of 3.61mm, 3.78mm and 10.25mm on WT, TC, ET. Compared to transformer-based methods, our method achieves significantly better performance with both metrics. Specifically, our approach outperforms TransBTS~\cite{wang2021transbts} and nnFormer~\cite{zhou2021nnformer} by more than 7\% and 4\% Dice on ET respectively. Besides, our method shows more competitive results than other SSL methods using the same image segmentation network. 
For ET category, our method obtains 3.46mm and 1.54mm lower in HD95 than 3D-RPL and 3D-Jig. The visual comparisons are shown in Fig. \ref{Visual_Seg}. Our method does predict the ET region (blue) more accurately.
As Table \ref{table_ibsr_wmh} shows, on IBSR dataset our method displays the highest Dice on CSF \& GM, and obtains the lowest HD95 on CSF. On WMH dataset, the proposed method performs the best on both metrics. These results show that the model weights pre-trained by our method can be transferred to a wide range of datasets and help achieve the state-of-the-art performance.

\subsubsection{Ablation Analysis.}
\begin{table}[!tbp]
    \centering
    \setlength{\tabcolsep}{1.3mm}{
    \caption{Comparison on IBSR dataset and WMH dataset.}
    \label{table_ibsr_wmh}
     \begin{tabular}{lcccccccc}
   \toprule
   Task & \multicolumn{6}{c}{IBSR} & \multicolumn{2}{c}{WMH} \\
   Metric  & \multicolumn{3}{c}{Dice(\%)$\uparrow$} &\multicolumn{3}{c}{HD95(mm)$\downarrow$} & \multicolumn{1}{c}{Dice(\%)$\uparrow$} &\multicolumn{1}{c}{HD95(mm)$\downarrow$} \\
   \cmidrule(lr){2-4} \cmidrule(lr){5-7} \cmidrule(lr){8-9} 
   Anatomy & CSF & GM & WM & CSF & GM & WM & WMH & WMH  \\
   \midrule[2pt]
   nnFormer \cite{zhou2021nnformer} & 87.31 & 93.81 & 92.12 & 1.52 & \textbf{1.52} & \textbf{1.21} & 78.04 & 2.81\\ 
   TransBTS \cite{wang2021transbts} & 81.42 & 93.91 & 92.17 & 7.84 & 1.54 & 1.40 & 78.81	& 2.91 \\
   UNETR \cite{hatamizadeh2022unetr} & 86.75 & 93.49 & 91.86 & 1.64 & 1.74 & 1.48 & 77.99 & 3.53 \\
   3D-RPL \cite{taleb20203d} & 86.63 & 93.85 & \textbf{92.50} & 1.83 & 1.54 & 1.29 & 78.63 & 3.06 \\
   3D-Jig \cite{taleb20203d} & 86.93 & 93.57 & 92.11 & 2.00 & 1.74 & 1.44 & 77.86 & 3.36 \\
   Ours & \textbf{87.63} & \textbf{93.91} & 92.44 & \textbf{1.46} & 1.54 & 1.33 & \textbf{78.99} & \textbf{2.73}\\
   \bottomrule
\end{tabular}}
\end{table}

\begin{table}[!h]
    \centering
    \setlength{\tabcolsep}{3mm}{
    \caption{Ablation study on BraTS 2021. SSL denotes the 3D Masked Autoencoder (MAE) SSL method. A-SSL is the MAE method with our AR-Loss.
    }
    \label{table_ablation}
    \begin{tabular}{lcccccc}
   \toprule
   Metric & \multicolumn{3}{c}{Dice(\%)$\uparrow$} &\multicolumn{3}{c}{HD95(mm)$\downarrow$}  \\
   \cmidrule(lr){2-4} \cmidrule(lr){5-7}
   Anatomy & WT & TC & ET & WT & TC & ET  \\
   \midrule[2pt]
   Baseline & 93.75 & 89.76 & 84.98 & 3.93 & 4.09 & 13.93 \\
   w/ SSL & 94.02 & 90.28 & 86.25 & 4.01 & 4.06 & 13.44\\
   w/ A-SSL & 93.95 & 90.24 & 86.38 & 3.84 & 3.79 & 11.69 \\
   \midrule[0.5pt]
   w/ SPE & 93.90 & 90.15 & 85.86 & 3.69 & 3.82 & 11.59\\
   w/ SPE\&SSL & 93.85 & 90.04 & \textbf{86.83} & 3.64 & 3.84 & 11.59 \\
   w/ ASA (Ours) & \textbf{94.03} & \textbf{90.29} & 86.76 & \textbf{3.61} & \textbf{3.78} & \textbf{10.25} \\
   \bottomrule
\end{tabular}}
\end{table}
We verify the strength of the attentive reconstuction loss (AR-Loss), the SPE, and our overall ASA framework on BraTS 2021, as shown in Table~\ref{table_ablation}. `Baseline' denotes the SW-ViT based segmentation network (see Fig.~\ref{Down-structure}) trained from scratch. `w/ SSL' denotes training the segmentation network with the model weights pre-trained by a 3D Masked Autoencoder (MAE) SSL method~\cite{he2021masked}. `A-SSL' denotes the MAE method with the proposed AR-Loss.
As shown as the first half of Table \ref{table_ablation}, the A-SSL method produces more accurate segmentation results than the competitor SSL at HD95 metric. Especially, on ET the HD95 of using A-SSL is nearly 2mm lower than that of using the SSL method. Note that HD95 measures the distance between the point sets of two boundaries. The above results show that the proposed AR-Loss can encourage the encoder to learn better representations for boundary information.
`w/ SPE' denotes applying the SPE to the train-from-scratch Baseline. `w/ SPE' obtains 0.9\% higher in Dice and 2.3mm lower in HD95 than `Baseline'. `w/ ASA' denotes using our overall method with the loss and SPE. By comparing ASA with A-SSL, the SPE can further slightly improve A-SSL by 1.4mm HD95 on ET. These results suggest that our proposed encoding can help the ViT-based encoder understand symmetric structures and harvest discriminative features.

\section{Conclusion}
In this paper, we propose a novel self-supervised learning architecture for 3D medical images. The proposed framework contains two key components, the symmetric position encoding and the attentive reconstruction loss.
The encoding can benefit feature learning for symmetric structures and the attentive loss emphasizes informative image regions for reconstruction-based SSL. Both techniques can improve the generalization of trained models.
Extensive experiments are conducted on three public brain MRI datasets. The results suggest that our method can achieve competitive performance with the state-of-the-art SSL methods and medical image segmentation models.

%
%
%
\bibliographystyle{splncs04}
\bibliography{mybibliography}

\end{document}